\documentclass[conference,letter]{IEEEtran}
\IEEEoverridecommandlockouts

\usepackage{color}
\usepackage{amsthm}
\usepackage{mathtools}

\usepackage{bbm}
\usepackage{amssymb}
\usepackage{tikz}
\usepackage{enumitem}
\usepackage[algo2e,ruled,vlined,linesnumbered]{algorithm2e}
\SetInd{0em}{0.5em}
\SetKwComment{Comment}{/\!\!/}{}
\usepackage{booktabs}
\usepackage{cite}
\usepackage{steinmetz}
\usepackage{subcaption}
\usepackage{dblfloatfix}

\newcommand\blfootnote[1]{%
  \begingroup
  \renewcommand\thefootnote{}\footnote{#1}%
  \addtocounter{footnote}{-1}%
  \endgroup
}

\usepackage{caption}
\newcommand{\citep}[1]{\cite{#1}}

\renewcommand{\hat}{\widehat}

\def\beq{\begin{equation}}
\def\eeq{\end{equation}}
\def\beqa{\begin{eqnarray}}
\def\eeqa{\end{eqnarray}}
\def\beqan{\begin{eqnarray*}}
\def\eeqan{\end{eqnarray*}}

\DeclareMathOperator*{\argmin}{arg\,min}

\theoremstyle{definition}

\def\Exp{\mathbb{E}}

\newcommand{\scan}{\mathrm{Scan}}
\newcommand{\beam}{\mathrm{Beam}}

\newcommand{\dbf}{\mathbf{d}}

\newcommand{\wbf}{\mathbf{w}}
\newcommand{\nbf}{\mathbf{n}}

\newcommand{\zbf}{\mathbf{z}}

\newcommand{\Hbf}{\mathbf{H}}

\newcommand{\herm}{^{\text{\sf H}}}

\begin{document}
\title{On Single-User Interactive Beam Alignment in Next Generation Systems: A Deep Learning Viewpoint}
\author{Abbas Khalili, Sundeep Rangan, Elza Erkip\\
 NYU Tandon School of Engineering,\\
Emails: \{ako274,srangan,elza\}@nyu.edu}

\maketitle

\begin{abstract}
Communication in high frequencies such as millimeter wave and terahertz suffer from high path-loss and intense shadowing which necessitates beamforming for reliable data transmission. On the other hand, at high frequencies the channels are sparse and consist of few spatial clusters. Therefore, \textit{beam alignment} (BA) strategies are used to find the direction of these channel clusters and adjust the width of the beam used for data transmission. In this work, a single-user uplink scenario where the channel has one dominant cluster is considered. It is assumed that the user transmits a set of BA packets over a fixed duration. Meanwhile, the base-station (BS) uses different probing beams to scan different angular regions. Since the BS measurements are noisy, it is not possible to find a narrow beam that includes the angle of arrival (AoA) of the user with probability one. Therefore, the BS allocates a narrow beam to the user which includes the AoA of the user with a predetermined error probability while minimizing the expected beamwidth of the allocated beam. Due to intractability of this noisy BA problem, here this problem is posed as an end-to-end optimization of a deep neural network (DNN) and effects of different loss functions are discussed and investigated. It is observed that the proposed DNN based BA, at high SNRs, achieves a performance close to that of the optimal BA when there is no-noise and for all SNRs, outperforms state-of-the-art.

\blfootnote{This work is supported by
National Science Foundation grants EARS-1547332, 
SpecEES-1824434, and NYU WIRELESS Industrial Affiliates.}
\end{abstract}

\section{Introduction}
Next generations of wireless networks (e.g., $5$G and $6$G) utilize high bandwidths available at millimeter wave (mmWave) and terahertz (THz) frequencies to achieve multi-Gbps throughputs \cite{mmWave-survey-nyu,rappaport2019wireless}. However, signals transmitted in these frequencies suffer from high path-loss and intense shadowing which are major obstacles towards practical realization of these systems. To mitigate the path-loss, several beamforming (BF) methods were proposed which use directional beams (a.k.a. \textit{narrow beams}) that concentrate the power towards the direction of interest\cite{kutty2016beamforming}.

A BF method should have low complexity for practical implementation and small overhead as the goal is to maximize the system's throughput. One can take advantage of the characteristics of the channel to design a suitable BF method. It is shown through experimental results that due to the properties of wave propagation at mmWave and THz frequencies, channels have few spatial clusters \cite{akdeniz2014millimeter,xing2018propagation}. Therefore, beam search methods are used to perform BF and to find a narrow beam for data transmission which is aligned with the the channel, i.e., the angle of arrival (AoA) and angle of departure (AoD) associated with channel clusters \cite{giordani2018tutorial}. These methods are also known as \textit{beam training} and \textit{beam alignment} (BA). BA techniques can be used for establishing connection between the users and the base station (BS) \cite{barati2016initial,giordani2016comparative}, or for beam tracking whose goal is to update the beam directions as the channel's AoA or AoD change due to the user's mobility or the propagation environment \cite{Shah1906:Robust}. It is also typically, assumed that the transceivers only use one-RF for BA due to high power consumption. This is known as performing analog BA. 

BA methods can be categorized into two main classes, namely \textit{interactive} and \textit{non-interactive}. In non-interactive BA, the transmitter sends a set of BA packets through a fixed set of probing beams and then allocates a data beam to the receiver based on the received feedback for the BA packets. In interactive BA methods, however, the probing beams for each BA packet can be modified based on the received feedback for the prior BA packets. As a result of this feedback, interactive BA methods usually have superior performance compared to non-interactive BA. 

Any BA problem can be thought of optimizing a policy function which determines the probing beams used at each time-slot to minimize a certain objective. Except for a few spacial cases, such as interactive and non-interactive BA with no measurement noise \cite{khalili2020optimal,khalili2021single,michelusi2018optimal}, this optimization is not tractable. On the other hand, deep neural networks (DNNs) have shown a great potential in performing non-tractable complex optimizations in various fields such as computer vision and natural language processing. Moreover, DNNs are  universal function approximators \cite{hornik1989multilayer} which makes them suitable for approximating the unknown functions such as BA policy function in our case. Therefore, DNNs  could be a suitable tool for handling complex BA problems. 

There is a large body of work on different BA techniques developed for different scenarios and objectives \cite{khalili2020optimal,khalili2021single,michelusi2018optimal,hussain2017throughput,barati2016initial,giordani2016comparative,Shah1906:Robust,khosravi2019efficient,chiu2019active,shabara2018linear,klautau20185g,Song2019}. More specifically, reference \cite{chiu2019active} considers noisy interactive analog BA where the objective is to find a data beam with a target resolution that includes the user AoA with a given error probability. They provide an a heuristic approach based on active learning that leads to the best known performance for the considered problem. Beam training using DNNs has also been investigated in variety of setups \cite{anton2019learning,ma2020machine,heng2019machine,wang2018mmwave,alkhateeb2018deep}. These works use DNNs to learn optimal BF vectors based on information on the physical environment.

In this paper, we consider a similar set up as of \cite{chiu2019active} and investigate single-user uplink interactive BA where the user has an omnidirectional transmission pattern while the BS performs analog BA and obtains noisy measurements.
Similar to \cite{chiu2019active}, we assume that the channel has one dominant cluster. Also, motivated by prior works \cite{barati2016initial,giordani2016comparative} and due to practical constraints, we consider that the BS can only use contiguous beams (i.e., beams with contiguous angular coverage regions). We also assume an arbitrary distribution on the user AoA to account for prior knowledge which could arise in user tracking applications \cite{Shah1906:Robust}.
At the end of the BA, the BS needs to allocate a narrow data beam to the user. Since the BS measurements are noisy, to find a narrow beam that includes the AoA we should consider errors in the beam including the AoA. Our objective is to minimize the expected beamwidth of the data beam allocated to the user subject to a given error probability. 
Unlike works \cite{anton2019learning,ma2020machine,heng2019machine,wang2018mmwave,alkhateeb2018deep} which generate a data set based on the environment physics and train DNNs to approximate a mapping for the optimal beams, we use a DNN to find a BA strategy based on AoA prior and the BS measurements. Compared to \cite{chiu2019active} where a fixed set of beams is considered and a policy for probing beam selection at each time-slot is given, we use a DNN to jointly optimize the set of probing beams and the probing beam selection policy.
Our main contributions are as follows:
\begin{itemize}
    \item We formulate the aforementioned BA problem  as an end-to-end optimization of a DNN using a recurrent neural network (RNN) and propose and discuss different loss functions for the considered objective (Section~\ref{sec:PS}).
    \item Through simulations, we observe that at high SNRs, the proposed solution performs close to the optimal BA, optimized for the case of no noise, in terms of minimum mean square error (MMSE) and expected beamwidth for uniform prior on AoA. Moreover, we show that the proposed DNN based BA outperforms state-of-the-art for all SNR regimes and different priors. Furthermore, we observe that for a given SNR and BA duration range, it is enough to train the network for only a fixed SNR and/or BA duration to get a performance close to optimized. This is important as it shows the framework is robust to not knowing the channel fading coefficient (Section~\ref{sec:sim}).
\end{itemize}

Recent concurrent work in \cite{sohrabi2020deep} also uses DNNs to optimize the BA procedure in \cite{chiu2019active}. However, unlike \cite{sohrabi2020deep} that only considers MMSE for AoA estimation, here we provide results for both MMSE of AoA estimation and minimum expected beamwidth along with proposing new loss functions, and analyze the robustness of the DNN framework to the BA duration and received signal SNR. Our emphasize on minimizing expected beamwidth is because of the importance of using narrow beams for data communication in mmWave and THz systems.
 
\section{System Model and Preliminaries} 
\label{sec:sys}
\subsection{Network Model}
\begin{figure}[t]
\centering
\includegraphics[width=0.5\linewidth, draft=false]{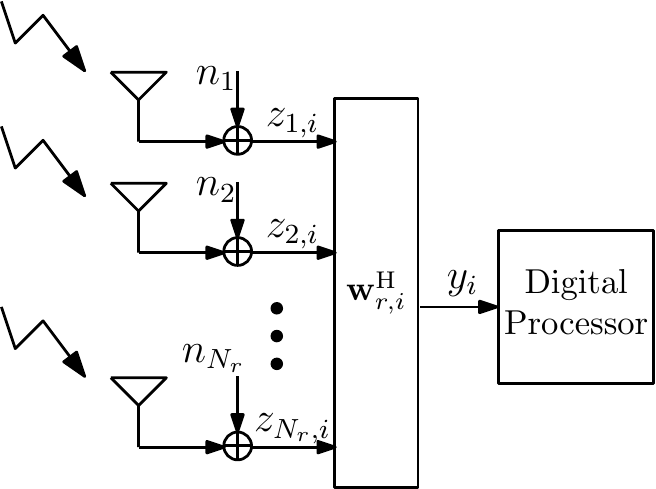}
\caption{ Receiver front end with analog beamforming.}
\label{fig:sys_model}
\vspace*{-0.4cm}
\end{figure}

We consider a single-user uplink communication in a single-cell scenario. We assume that the user performs omnidirectional transmission and sends a BA symbol in each time-slot while the BS performs analog BA and scans an angular interval. 
Let us denote the transmitted BA symbol from the user by $s$ where $|s|^2 = 1$, then the received beamformed signal at the BS, at the $i^{\rm th}$ time-slot is
\begin{align}
\label{eq:inout}
    y_i = \wbf_{r,i} \herm \zbf_i =  \sqrt{P} \wbf_{r,i} \herm \Hbf \wbf_t s + \wbf_{r,i} \herm \nbf_i,
\end{align}
where $\zbf_i$ is the received signal prior to beamfomring, $\wbf_t$ and $\wbf_{r,i}$ are the beamforming vectors at the user and BS with $|\wbf_{r,i}|^2 =|\wbf_{t}|^2 =1$, $\Hbf$ is the gain matrix of the channel between the user and BS, and $\nbf_i, i\in[b]$ is the additive Gaussian noise with i.i.d. $\mathcal{CN}(0,\sigma^2)$ elements\footnote{We denote by $[b]$ the set of integers from 1 to $b$.}. A detailed illustration of the receiver front-end is provided in Fig.~\ref{fig:sys_model}.

\begin{figure*}[t]
\centering
\includegraphics[ width=0.5\linewidth, draft=false]{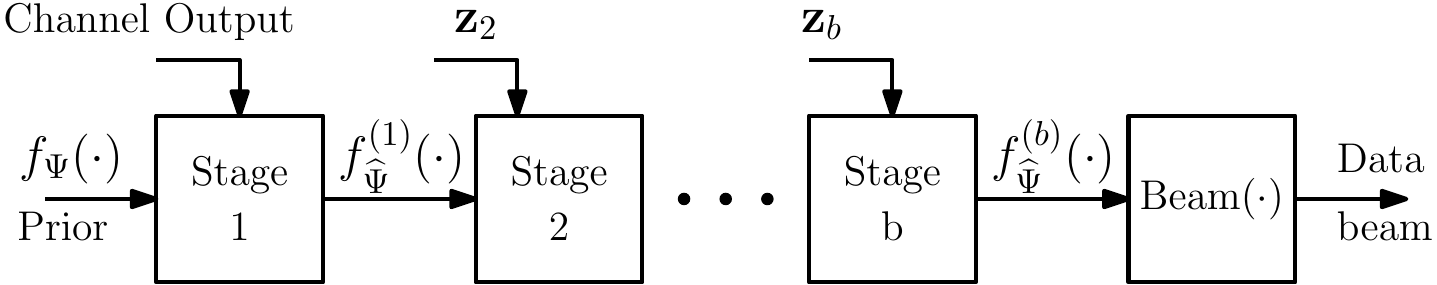}
\caption{The block diagram of the BA phase.}
\label{fig:RNN}
\vspace*{-0.4cm}
\end{figure*}

\begin{figure}[t]
\centering
\includegraphics[ width=0.7\linewidth, draft=false]{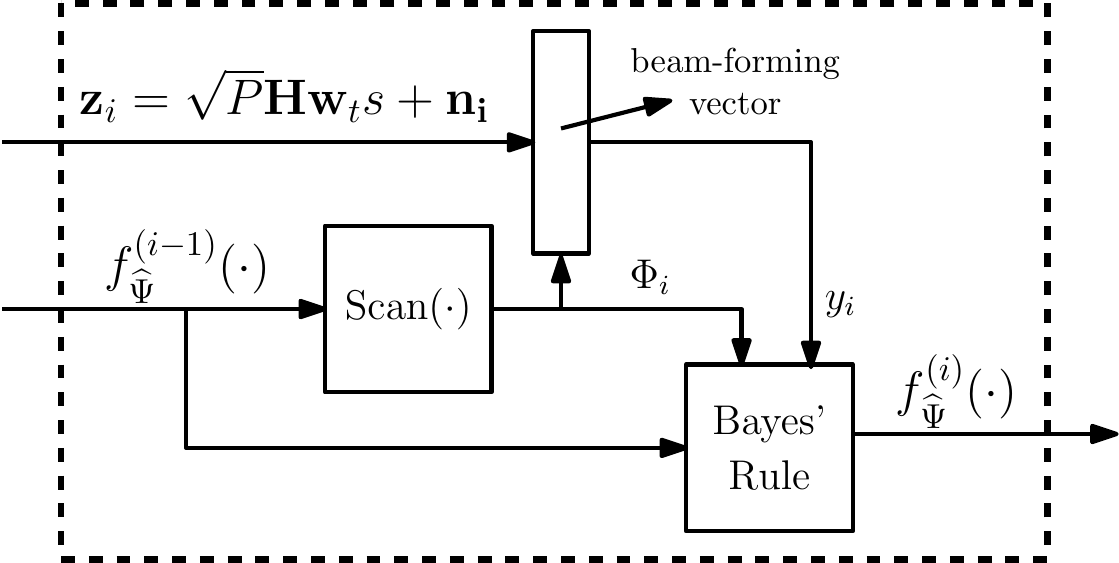}
\caption{BA details operation at stage $i$ of Fig.~\ref{fig:RNN}.}
\label{fig:stage}
\vspace*{-0.4cm}
\end{figure}

Supported by experimental studies \cite{akdeniz2014millimeter,rappaport2019wireless,nguyen2018comparing} and similar to works \cite{hussain2017throughput,michelusi2018optimal}, we assume that the propagation channel between the BS and the user consists of one dominant path (spatial cluster). Using clustered channel model \cite{akdeniz2014millimeter}, we have 
\begin{align}
\label{eq:cch}
    \Hbf  = \sqrt{N_t N_r}h\dbf_r(\psi_r)\dbf_t\herm(\psi_t),
\end{align}
where $N_t$ and $N_r$ denote the number of transmit and receive antennas, respectively, $\psi_t$ and $\psi_r$ represent the AoD and AoA of the channel cluster, respectively, $h$ is the channel gain including path loss, and the terms $\dbf_t(\cdot)$ and $\dbf_r(\cdot)$ are the array response vectors of the user and BS, respectively. We assume that the channel is stationary in the time interval of interest meaning that the spatial cluster is fixed. Furthermore, for the purpose of the analysis, we assume that $h$ is known. However, we will show through simulations in Sec.~\ref{sec:sim} that not knowing $h$ would not affect the performance significantly. Combining \eqref{eq:inout} and \eqref{eq:cch}, we get 
\begin{align}
\label{eq:yi}
    y_i  = h\sqrt{P G(\wbf_{r,i},\psi_r)G(\wbf_t,\psi_t)}e^{j \gamma_i(\psi_r, \psi_t)}s + \hat{\nbf}_i,
\end{align}
where $ G_{x}(\wbf_x,\psi_x) = N_x|\dbf_x \herm \wbf_x|^2, x\in\{t,\{r,i\}\}$ are beamforming gains at the transmitter and receiver, respectively, $\gamma_i(\psi_r, \psi_t) = \angle\dbf_r\herm \wbf_{r,i} -\angle\dbf_t\herm \wbf_t $ is a phase component, and $\hat{\nbf}_i = \wbf_{r,i} \herm \nbf $.
The omnidirectional transmission at the user yields to $G_{t}(\wbf_t,\psi_t) = 1$. Based on \eqref{eq:yi}, we define \textit{raw SNR} as the averaged received SNR per antenna at the BS prior to beamforming. We have the raw SNR as $\frac{h^2P}{\sigma^2}$.

Motivated by prior works \cite{khalili2020optimal,michelusi2018optimal} we only consider the case where the beams are contiguous. For our analysis, we assume that the BS has a massive antenna array as envisioned for the next generations of wireless systems \cite{mmWave-survey-nyu}. Therefore, we can adopt the \textit{sectored antenna} model from \cite{ramanathan2001performance} which has been used in several other studies to model the BF gain of antenna arrays (e.g. \cite{bai2015coverage,khalili2020optimal}). In this model each beam $\Phi$ is characterized by two parameters: a constant main-lobe gain and the angular coverage region (ACR) which is the angular interval covered by the main-lobe. As a result of this model, we have
\begin{align}
    G(\wbf_{r,i},\psi_r) \approx \frac{2\pi}{|\Phi_{r,i}|} \mathbbm{1}_{\{\psi_r \in \Phi_{r,i}\}}
\end{align}
where $\Phi_r$ is the beam associated with the beamforming vector $\wbf_r$ and $\mathbbm{1}_{\{\mathcal{A}\}}$ is the indicator function of event $\mathcal{A}$. From here onward, we drop the sub-index $r$ for readability. 

\begin{figure*}[t]
\centering
\begin{subfigure}{0.31\linewidth}
    \centering
    \includegraphics[width=\textwidth]{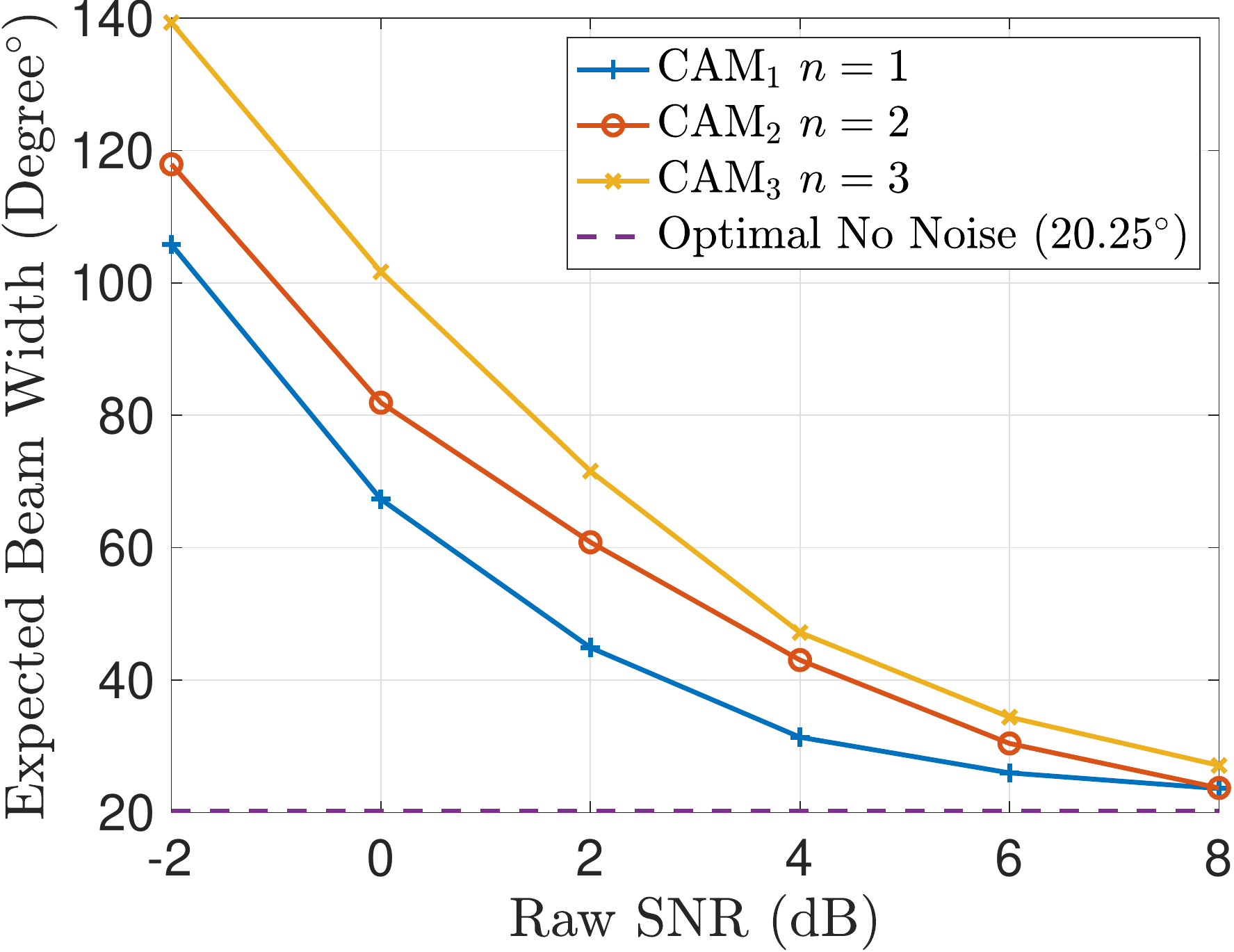}
    \caption{}
    \label{fig:loss_meanl}
\end{subfigure}
\begin{subfigure}{0.31\linewidth}
    \centering
    \includegraphics[width=\textwidth]{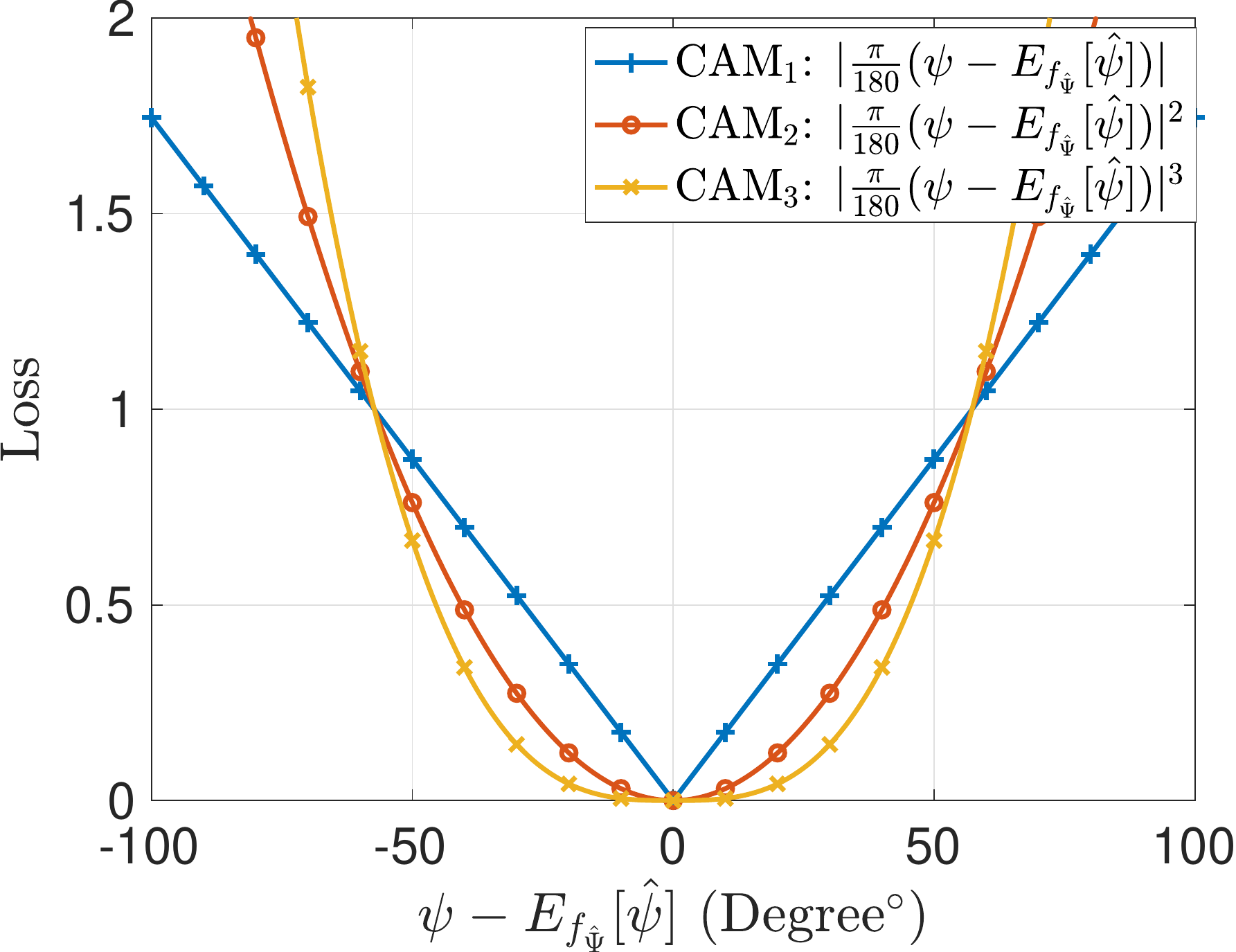}
    \caption{}
    \label{fig:loss_norm}
\end{subfigure}
\begin{subfigure}{0.31\linewidth}
    \centering
    \includegraphics[width=\textwidth]{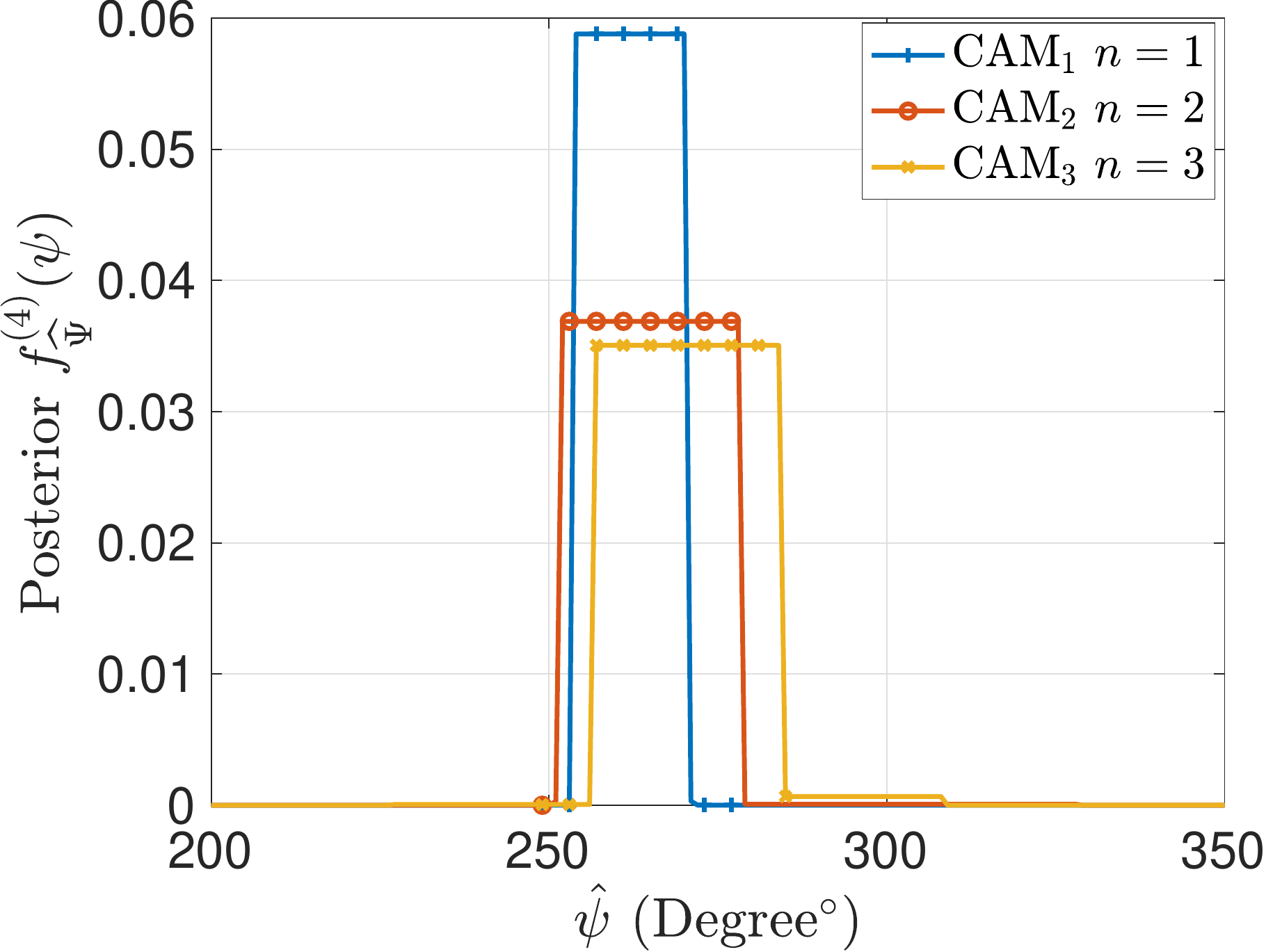}
    \caption{}
    \label{fig:post}
\end{subfigure}
\vspace*{-0.1cm}
\caption{Performance comparison of different orders $n\in\{1,2,3\}$ of CAM loss function for $b=4$ and $\epsilon=0.1$. (a) Expected beamwidth for CAM. (b) CAM loss before taking the expectation with respect to $\Psi$. (c) Resulting posteriors for realized AoA$=\frac{3\pi}{2}$ and raw SNR$=8$ (dB).}
\vspace*{-0.4cm}
\end{figure*}

\subsection{Problem Formulation}
We assume that the user transmits a total of $b$ BA symbols each in one time-slot and the BS uses $b$ probing beams with ACRs $\{\Phi_i\}_{i\in[b]}$ to scan the angular space. 
Afterwards, the BS allocates a beam with ACR $\beam(\Psi)$ to the user for data transmission. Since the BS measurements are noisy, we require that the data beam should include the user AoA with probability $1-\epsilon$, for a given error probability $\epsilon$. We also assume that a prior $f_{\Psi}$ on the AoA is given. Our objective is, for a given $b$, $\epsilon$, and $f_{\Psi}$, to find a BA policy function $\scan(\cdot)$ that determines the probing beams at each time-slot to minimize the expected length of $\beam(\Psi)$ while ensuring the desired error probability. This problem can be written as the following optimization
\begin{equation*}
\scan^*(\cdot)= \argmin_{\scan(\cdot)} \mathbb{E}_{f_{\Psi}}[|\beam(\Psi)|],
\end{equation*}
where the expectation is with respect to random variable $\Psi$. 
Although there are closed-form solutions to this optimization in some special cases such as uniform prior with no noise where optimal solution becomes \textit{bisection method}, this optimization is not tractable in general. In next section, we will show how this problem can be posed as an end-to-end optimization of a DNN.

\section{Proposed Solution} 
\label{sec:PS}

Similar to \cite{hussain2017throughput} and \cite{michelusi2018optimal}, we view the BA problem as a Markov decision process  \cite{bertsekas1995dynamic} over the time interval $b$ as illustrated in Fig.~\ref{fig:RNN}. The state at the start of the stage $i$ is $(f^{(i-1)}_{\hat{\Psi}}(\hat{\psi}, \Psi),{y}_i)$, where $f^{(i-1)}_{\hat{\Psi}}(\hat{\psi}, \Psi)$ is the probability distribution over the AoA at the $(i-1)^{\rm th}$ time-slot and $f^{(0)}_{\hat{\Psi}}(\hat{\psi}, \Psi) = f_{\Psi}(\hat{\psi})$. Note that the distribution $f^{(i)}_{\hat\Psi}$ is a function of channel measurements $y_1, \ldots, y_i$ which depend on the realization of $\Psi$ and the prior on $\Psi$.

At each stage $i$, the posterior $f^{(i)}_{\hat{\Psi}}(\hat{\psi}, \Psi)$ can be calculated using Bayes' rule from the prior  $f^{(i-1)}_{\hat{\Psi}}(\hat{\psi}, \Psi)$, the function $\scan(\cdot)$, the resulting probing beam $\Phi_i$, and signal $y_i$. The update rule is
\begin{align}
    &f^{(i)}_{\hat{\Psi}}(\hat{\psi}, \Psi) =\nonumber\\
    &\frac{f^{(i-1)}_{\hat{\Psi}}(\hat{\psi}, \Psi) f_{Y}(y_i|\Psi = \hat{\psi}, \scan(f^{(i-1)}_{\hat\Psi}) = \Phi_i)}{\int_{\hat\psi} f^{(i-1)}_{\hat{\Psi}}(\hat{\psi}, \Psi) f_{Y}(y_i|\Psi = \hat \psi, \scan(f^{(i-1)}_{\hat\Psi}) = \Phi_i))d\hat\psi},
\end{align}
where $f_{Y}(y_i|\Psi = \hat\psi, \scan(f^{(i-1)}_{\hat\Psi}) = \Phi_i)$ is the conditional distribution of the received beamformed signal $y$ given that the true AoA is the angle $\hat\psi$ and the probing beam used at the $i^{th}$ time-slot is $\Phi_i$. Based on the channel model \eqref{eq:yi}, 
\begin{align}
     &f_{Y}(y_i|\Psi = \hat\psi, \scan(f^{(i-1)}_{\hat\Psi}) = \Phi_i) =\nonumber\\ &\qquad\qquad\qquad\quad \mathcal{CN}\left(y_i; h\sqrt{P G(\Phi_i,\hat\psi)}e^{j \gamma(\Phi_i)}s, \sigma^2 \right).
\end{align}

Consider Fig.~\ref{fig:stage} which shows the operations of the BS at the $i^{\rm th}$ stage (time-slot) of BA. The function $\scan(\cdot)$ is our optimization variable. 
It is known that DNNs are universal approximators of functions. Therefore, we can approximate the function $\scan(\cdot)$ using a neural network which is concatenation of a set of linear and non-linear layers. If we replace $\scan(\cdot)$ block in Fig.~\ref{fig:stage} with a neural network, the block diagram in Fig.~\ref{fig:RNN} would correspond to an RNN which can be optimized from end-to-end.

RNNs are neural sequence models which are typically used for time-series prediction as they can capture the dependencies of the elements in a time-series. In our problem the received signals are also dependent through the AoA; however, our task is different. Instead of a time-series prediction, we use an RNN to optimize and find the optimal BA policy function $\scan(\cdot)$ which minimizes the expected beamwidth of the data beam. 

To use a neural network to represent the function $\scan(\cdot)$, we need to discretize its input, which in this case, is a probability distribution. For resolution $2\pi/N$ in which the interval $(0, 2\pi]$ is divided into $N$ contiguous intervals with equal lengths, we use the network with the parameters presented in Table~\ref{tab:rnn}. At each time-slot, our proposed network is fed a prior and outputs two values which after being scaled by $2\pi$ represent the start angle and length of probing beam used at that time-slot.  

Another important component which needs to be determined in order to train the RNN is the loss function. Our objective is to minimize the expected beamwidth of the beam allocated to the user subject to error probability $\epsilon$. To this end, the BS uses the final posterior $f^{(b)}_{\hat\Psi}(\hat\psi, \Psi)$ to find the shortest contiguous beam which includes the AoA of the user with probability $1-\epsilon$. For a given realization $\psi$, this beam is the solution to the following optimization problem 
\begin{align}
    &\beam(\psi) = (\theta^*,\theta^*+\ell^*],\\
    &\begin{aligned}
    \label{eq:optbeam}
    (\theta^*,\ell^*) &= \argmin_{(\theta, \ell) \in{(0,2\pi]}^2}  \ell\\
    \textrm{such that}&~\int_\theta^{\theta+\ell} f_{\hat\Psi}^{(b)}(\hat\psi,\psi) d\hat\psi > 1-\epsilon.
    \end{aligned}
\end{align}
However, this is a constrained optimization without a closed-form solution and therefore, cannot be directly used to optimize the neural network. Here, we instead, propose the following loss functions.

\textit{Minimum Mean Square Error (MMSE) Loss:}
Consider a beamwidth $\ell$ satisfying the following:
\begin{align}
    \ell^2 = \frac{4}{\epsilon}\Exp_{f_{\Psi}}\left|\Psi-\Exp_{f^{(b)}_{\hat{\Psi}}}[\hat{\Psi}]\right|^2.
\end{align}
Using Markov's inequality, one can show that there always exists a beam of length $\ell$ for any realization of the AoA $\psi$ which includes the AoA with probability $\epsilon$.
Therefore, minimizing $\ell$ corresponds to minimizing the MMSE estimate of $\Psi$,  
\begin{align}
    \mathrm{MMSE} = \Exp_{f_{\Psi}}\left|\Psi-\Exp_{f^{(b)}_{\hat{\Psi}}}[\hat{\Psi}]\right|^2.
\end{align}
However, since the Markov's inequality maybe loose, the MMSE based optimization may not lead to the smallest expected beamwidth.

\textit{Central Absolute Moment Loss:} From \eqref{eq:optbeam}, we observe that the more concentrated the posterior $f_{\hat\Psi}^{(b)}(\cdot)$ is, the lesser the expected beamwidth should be. To enforce this concentration, we suggest minimizing the expected $n^{\rm th}$ central absolute moment (CAM) of the output posterior $f_{\hat\Psi}^{(b)}(\cdot)$ which is 
\begin{align*}
   \mathrm{CAM}_n =  \Exp_{f_{\Psi}} \left[\Exp_{f^{(b)}_{\hat{\Psi}}}\left| \hat{\Psi} - \Exp_{f^{(b)}_{\hat{\Psi}}}[\hat{\Psi}]\right|^n\right],
\end{align*}
where $n \in \mathbb{N}$ is the order of the moment whose optimal value is investigated in the next section. Note that the outer expectation emphasizes more likely AoA's and ensures they have lower beamwidths.

We will investigate the effects of these loss functions in more detail in the next section.
\begin{figure*}[t]
\centering
\begin{subfigure}{0.31\linewidth}
    \centering
    \includegraphics[width=\textwidth]{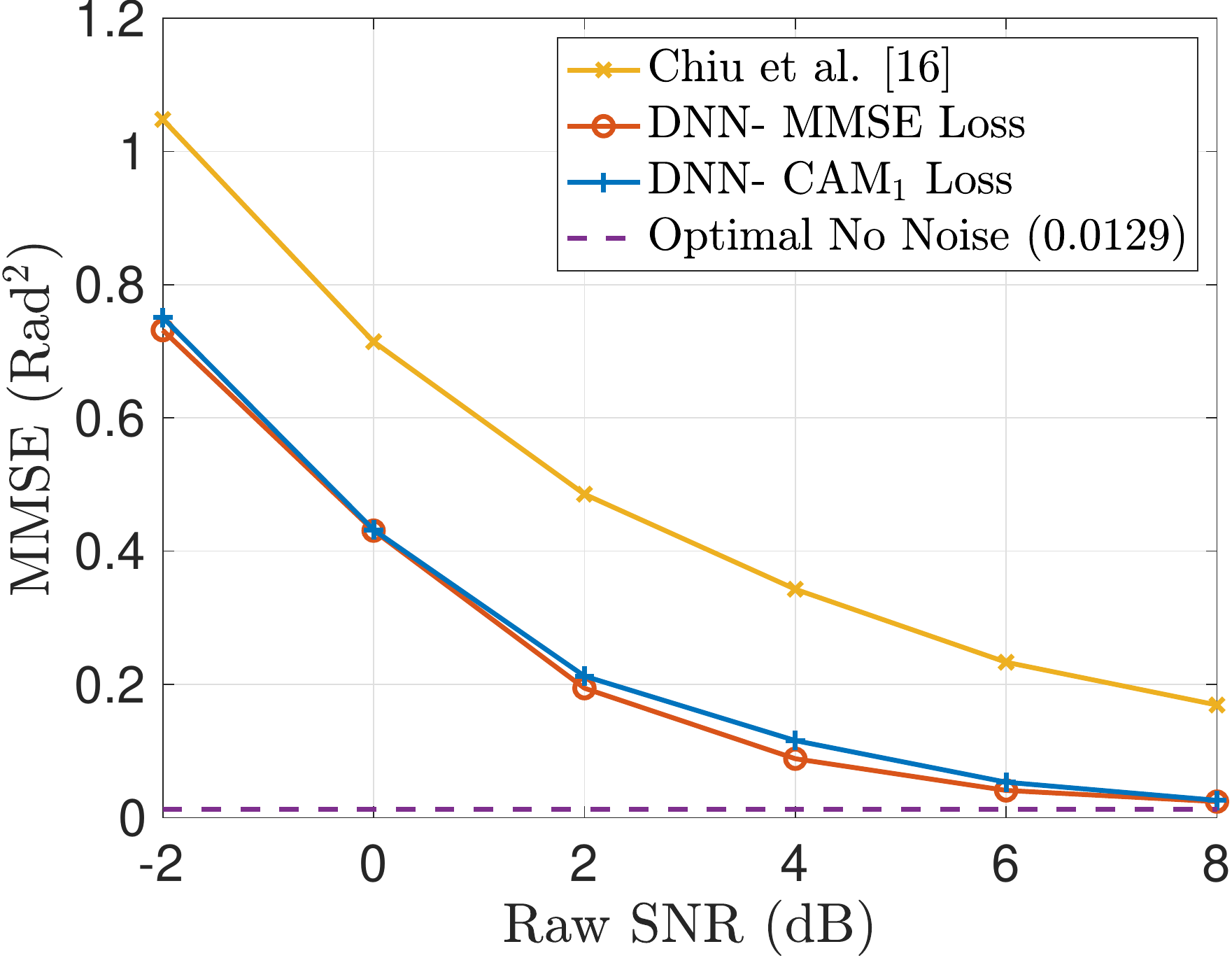}
    \caption{}
    \label{fig:mmse}
\end{subfigure}
\begin{subfigure}{0.31\linewidth}
    \centering
    \includegraphics[width=\textwidth]{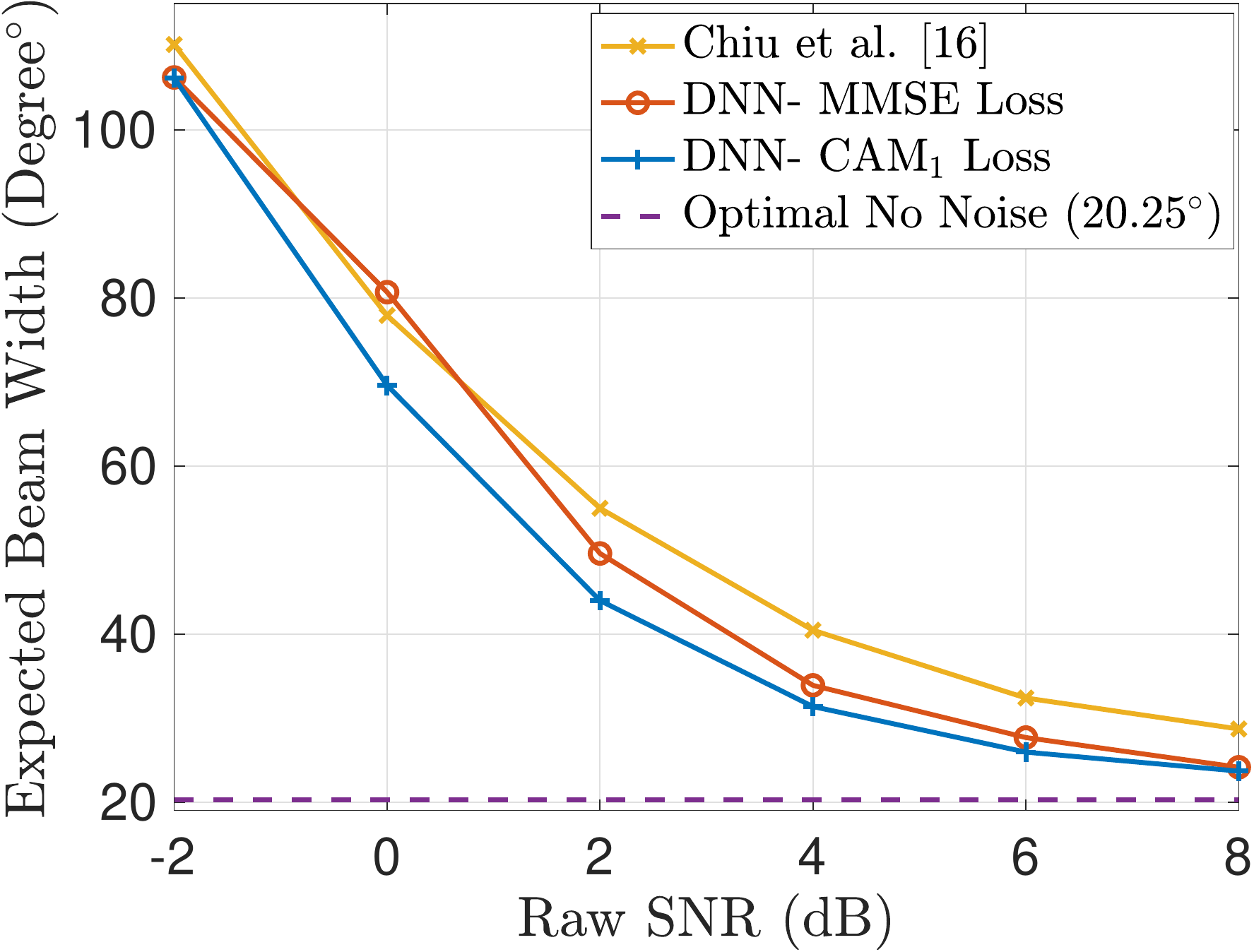}
    \caption{}
    \label{fig:meanl}
\end{subfigure}
\begin{subfigure}{0.31\linewidth}
    \centering
    \includegraphics[width=\textwidth]{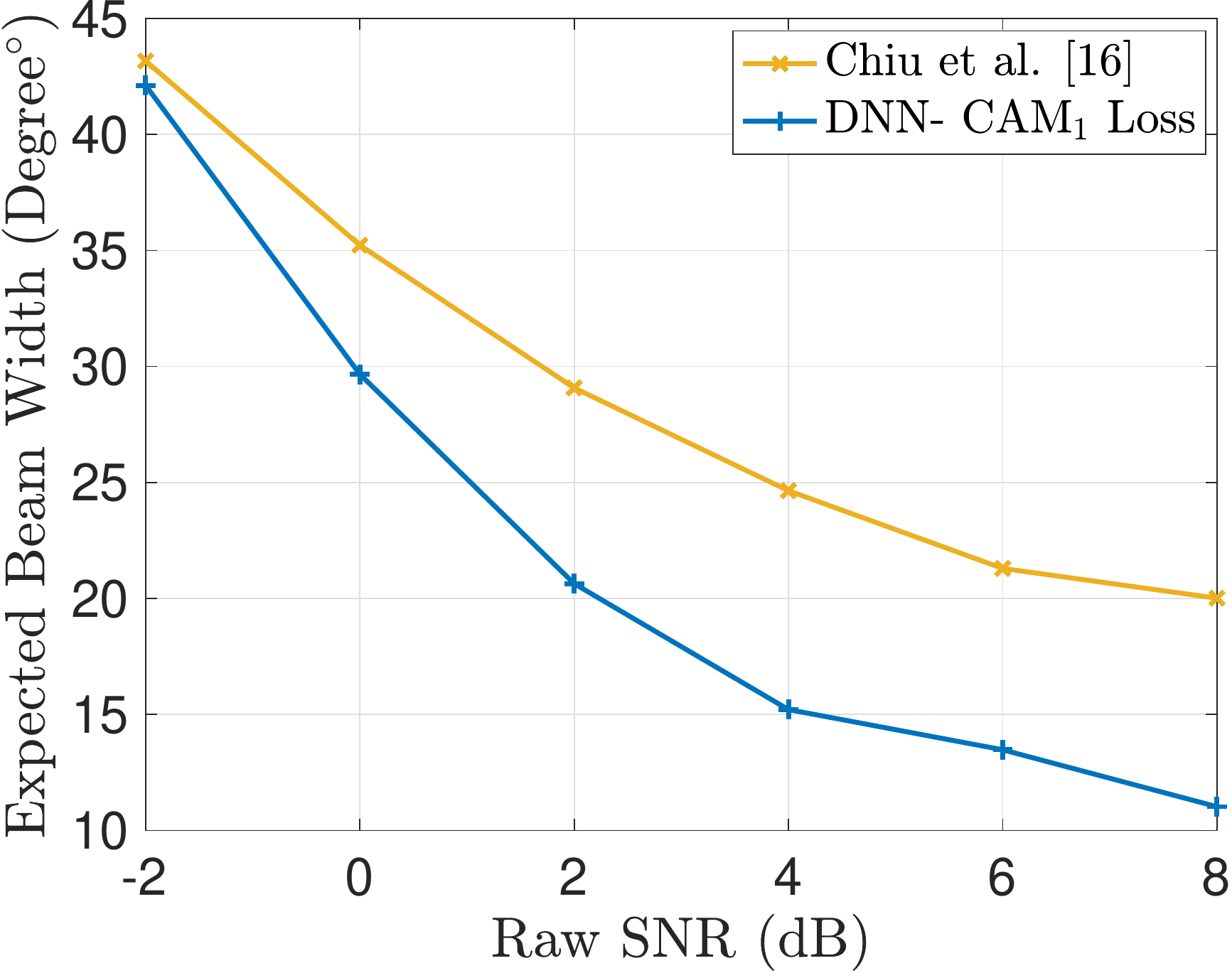}
    \caption{}
    \label{fig:meanl_pu}
\end{subfigure}
\vspace*{-0.1cm}
\caption{ Expected MMSE and beamwidth for different loss functions, $b=4$, $\epsilon = 0.1$, and different raw SNRs (a) Expected MMSE when $\Psi \sim \mathrm{Uniform}(0,2\pi]$. (b) Expected beamwidth when $\Psi \sim \mathrm{Uniform}(0,2\pi]$. (c) Expected beamwidth when $\Psi \sim \mathrm{Uniform}(5\pi/6,7\pi/6]$ with probability $0.9$ and uniform on the rest of the $(0,2\pi]$ with probability $0.1$.}
\vspace*{-0.4cm}
\end{figure*}

\begin{figure*}[b]
\vspace*{-0.4cm}
\centering
\begin{subfigure}{0.31\linewidth}
    \centering
    \includegraphics[width=\textwidth]{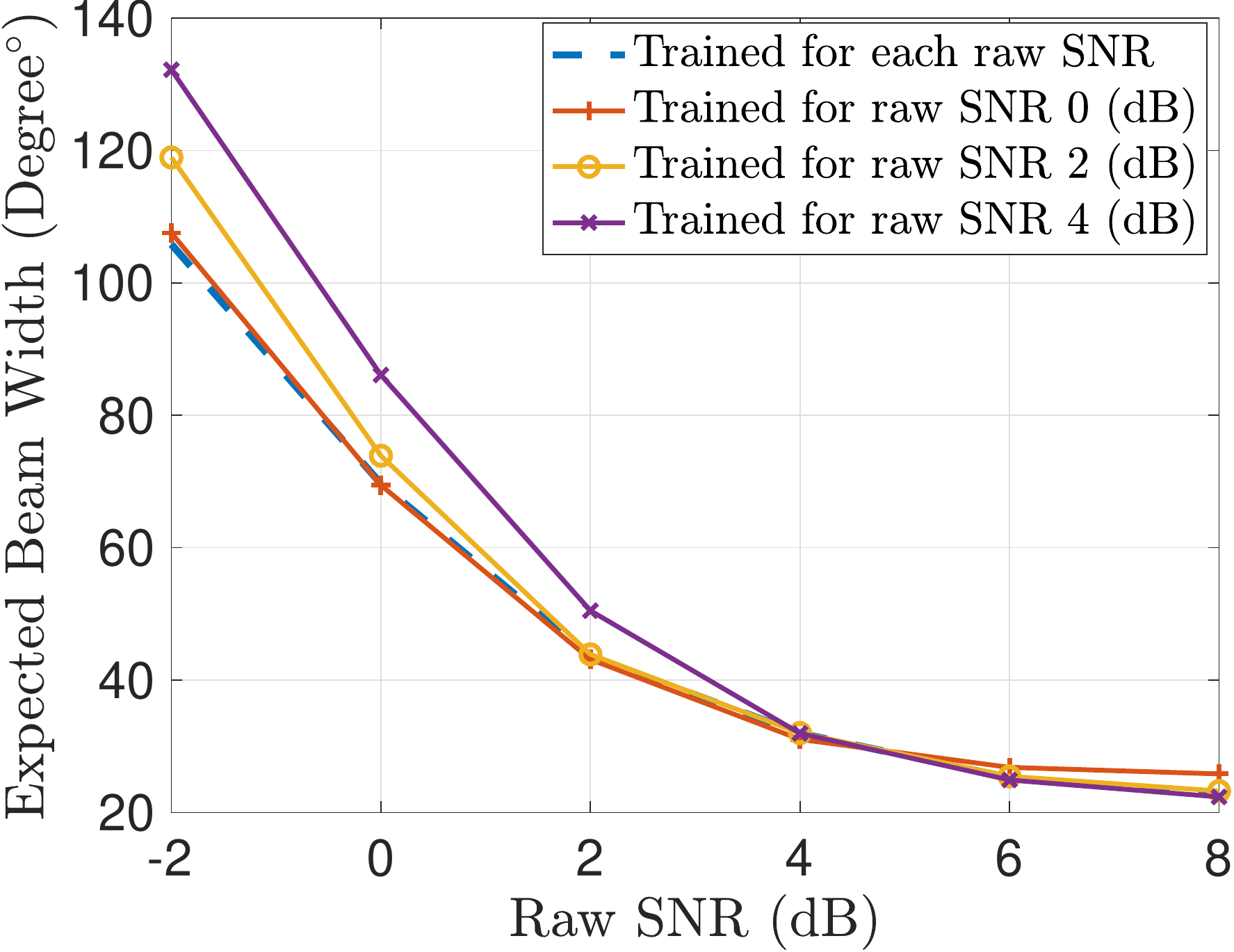}
    \caption{}
    \label{fig:gen_snr}
\end{subfigure}
\begin{subfigure}{0.31\linewidth}
    \centering
    \includegraphics[width=\textwidth]{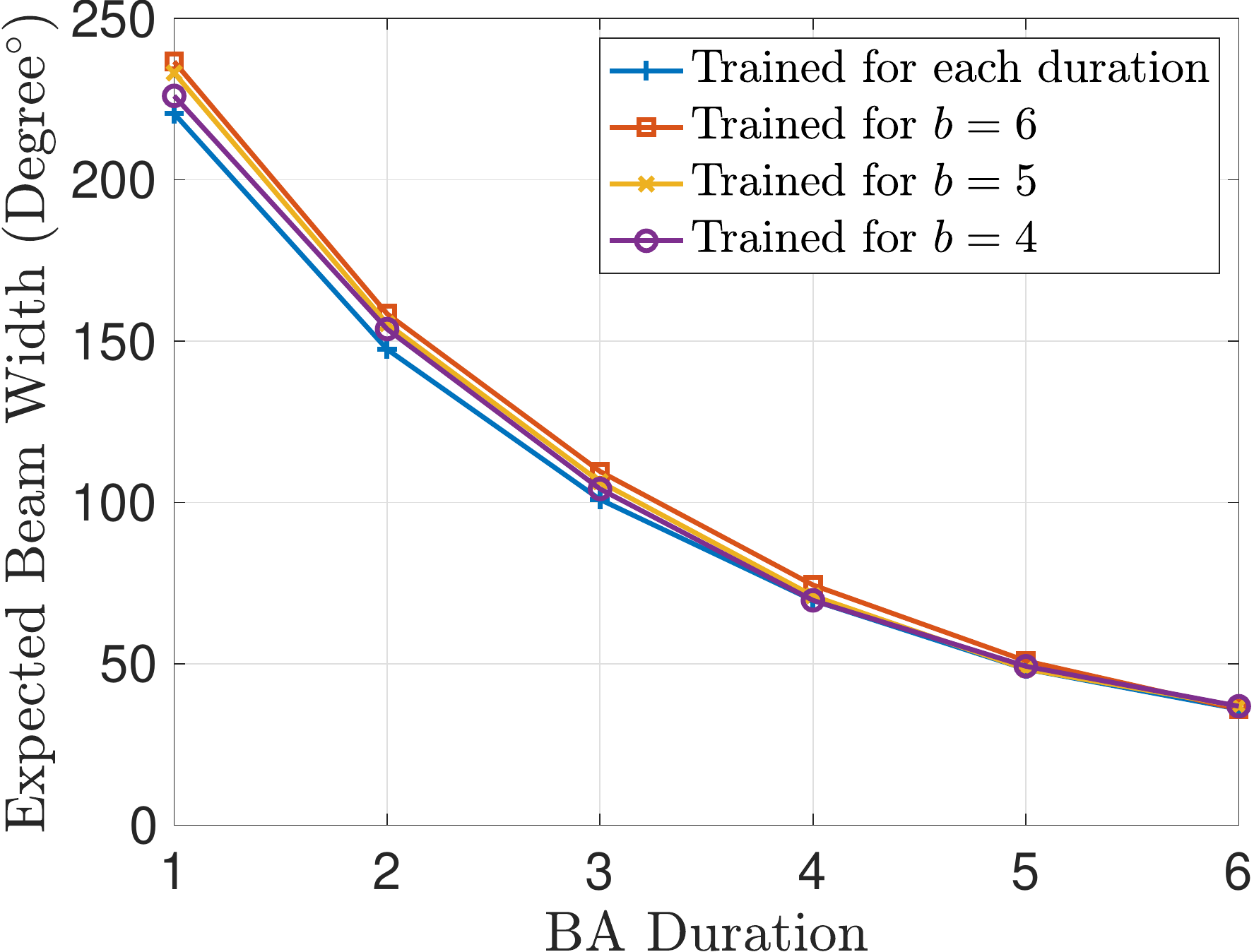}
    \caption{}
    \label{fig:gen_ba}
\end{subfigure}
\begin{subfigure}{0.31\linewidth}
    \centering
    \includegraphics[width=\textwidth]{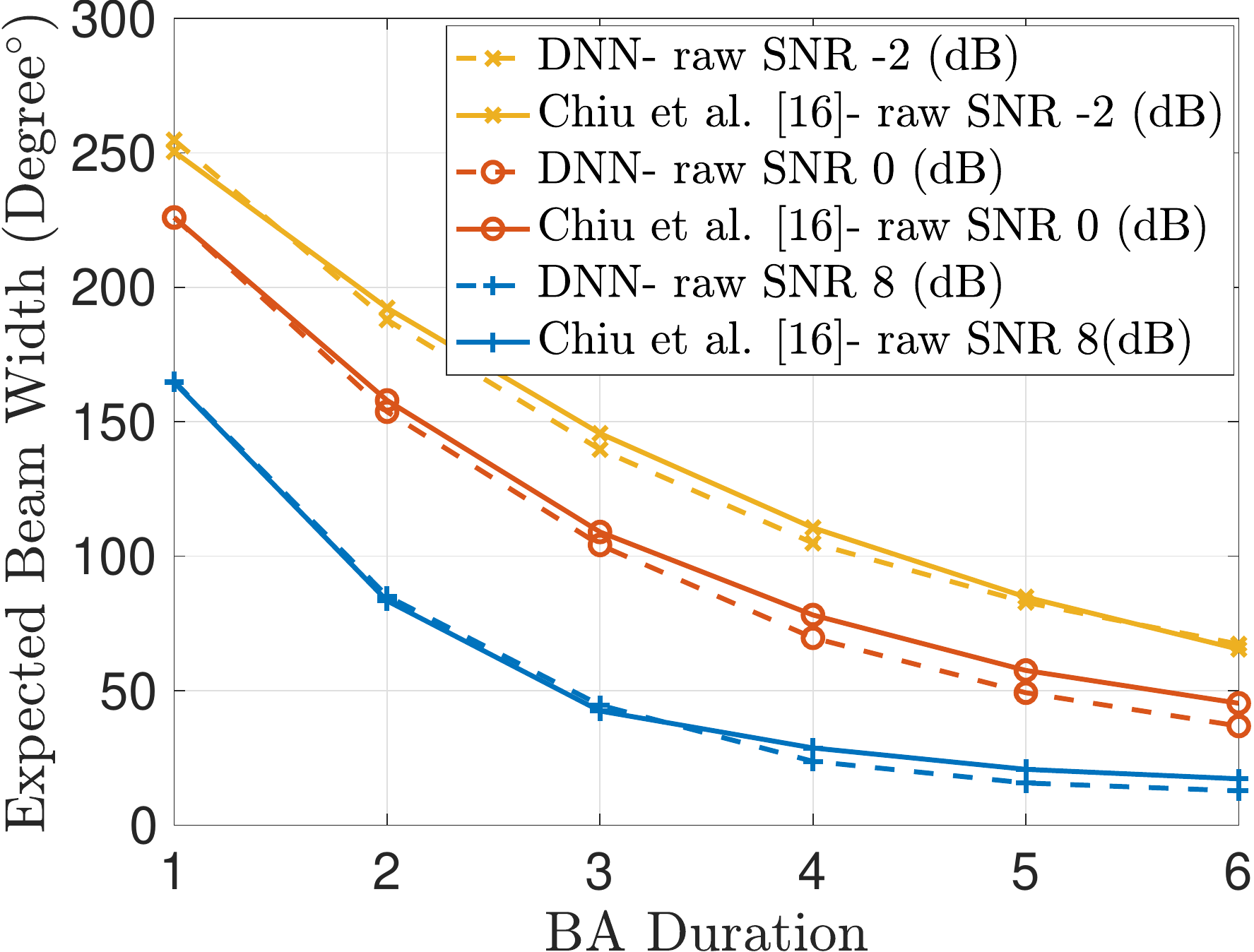}
    \caption{}
    \label{fig:fb}
\end{subfigure}
\vspace*{-0.1cm}
\caption{(a) Expected beamwidth for $b=4$ and different raw SNRs, when the network is trained for each raw SNR separately and when it is only trained for certain raw SNRs. (b) Expected beamwidth for raw SNR$=0$ (dB) and different BA durations when the network is trained for each duration separately and when it is only trained for only a fixed duration. (c) Expected beamwidth for different raw SNRs and BA durations when the network is trained for each raw SNR separately but only for duration $b=4$.}
\end{figure*}
\begin{table}[t]
\renewcommand{\arraystretch}{1.2} 
\centering
\caption{Layout of the Neural network  $\scan(\cdot)$ used in Fig.~\ref{fig:stage}
 It has ${12N^2 +10N}$ trainable parameters.
}
\label{tab:rnn}
\begin{tabular}{l|c}
 Layer    & Output dimensions    \\\hline
 Input & $N$ \\
 Dense + ReLU & $4N$ \\
 Dense + ReLU & $2N$ \\
 Dense + Sigmoid & $2$
\end{tabular}
\vspace*{-0.4cm}
\end{table}

\section{Simulations}
\label{sec:sim}
For our simulations, we discretize all pdfs with resolution of $1^\circ$ and divide the angular interval $(0,2\pi]$ into $N = 360$ equal length intervals. We consider the error probability of $\epsilon = 0.1$ and average the plots over $100000$ random realizations of AoA. Also, after training the RNN, to find the optimal data beam we perform the optimization in \eqref{eq:optbeam} using exhaustive search. We compare the performance of our proposed approach with that of \cite{chiu2019active} which has similar setup as our work.

We start by studying the performances of different orders $n$ of the CAM loss function. Let us consider the BA duration $b=4$. Figure~\ref{fig:loss_meanl} shows the expected beamwidth resulting from $n\in\{1,2,3\}$. We observe that the first order CAM results in the least expected beamwidth among the three. This can be justified using Fig.~\ref{fig:loss_norm} which shows the shape of the loss functions corresponding to different $\mathrm{CAM}$ orders. Compared to the cases $n=2$ and $n=3$, the case $n = 1$ has higher loss for the angles near the center of the distribution. In fact, this loss is a decreasing function of $n \in \mathbb{N}$. As a result, we would expect that the network lead to a more concentrated posterior for smaller values of $n$ which in turn leads to smaller expected beamwidth.  Fig.~\ref{fig:post} shows the resulted posteriors for different values of $n$ when realized $\psi = \frac{3\pi}{2}$ and raw SNR$=8$ (dB). It shows that as the order of CAM loss function is increased the resulting posterior becomes less and less concentrated which is consistent with Fig.~\ref{fig:loss_norm}. 
Based on these simulations, from here onward, we only consider the first order ($n=1$) CAM loss function.

Figure~\ref{fig:mmse} and Fig.~\ref{fig:meanl} show the expected MMSE and beamwidth of our proposed DNN based BA using the MMSE and $\mathrm{CAM}_1$ loss functions and the algorithm in \cite{chiu2019active}, respectively for different raw SNRs, $b=4$ and $\Psi \sim \mathrm{Uniform}(0,2\pi]$. As expected, we observe that MMSE loss function leads to the least mean square error compared to others, and at high raw SNRs, the performance gets close to the optimal MMSE when there is no noise. The optimal MMSE for the case of no noise can be derived using bisection algorithm and is equal to $\frac{\phi^2}{3\times2^{2b}} = 0.0129$\cite{Alkhateeb2014}. However, the $\mathrm{CAM}_1$ loss function outperforms the MMSE loss function in terms of  expected beamwidth. The rational is that, there can be multiple posteriors with same mean (resulting in same MMSE) but with different concentrations (leading to different expected beamwidths). Note that Fig.~\ref{fig:loss_meanl} also suggests that at high raw SNRs, the $\mathrm{CAM}_1$ loss function performs close to the optimal performance for the case with no noise which can be derived (using bisection algorithm) as $\frac{2\pi}{2^b}\times (1-\epsilon)$\cite{Alkhateeb2014}. For the rest of the simulations, we focus only on the expected beamwidth as it is the primary focus of the paper and only consider the CAM loss function as it has the least expected beamwidth. 

Figure~\ref{fig:meanl_pu} shows the expected beamwidth for the CAM loss function and the proposed algorithm in \cite{chiu2019active} for different raw SNRs and $b=4$ when $\Psi \sim \mathrm{Uniform}(5\pi/6,7\pi/6]$ with probability $0.9$ and uniform on the rest of the $(0,2\pi]$ with probability $0.1$. The result shows that our proposed method also out performs the one in \cite{chiu2019active} for a non-uniform prior. In this case, the difference between the expected beamwidths of the two methods are $(1^\circ,5.5^\circ,8.4^\circ, 9.4^\circ,7.8^\circ,9^\circ)$ for raw SNRs $\{-2,0,2,4,6,8\}$ (dB). This difference for the case of uniform prior in Fig.~\ref{fig:meanl} is  $(4^\circ,8.3^\circ,11^\circ, 9^\circ,6.4^\circ,5^\circ)$. Overall, we observe a higher difference at high raw SNRs for the case of non-uniform prior compared to the uniform prior. The reason is that the method in \cite{chiu2019active} uses a fixed hierarchical beam codebook which is optimal for uniform prior under no noise but its optimality for non-uniform priors is not guaranteed. The advantage of our solution is that the DNN has the ability to learn the optimal beams based on the prior which should allow for closer performance to the optimal compared to the method in \cite{chiu2019active}.

Finally, we investigate the ability of the DNN to generalize. Figure~\ref{fig:gen_snr} considers the case $b=4$ and illustrates the expected beamwidth for different raw SNRs when the network is trained for each raw SNR separately and when the network is trained only for specific raw SNRs. We see that by training the network for only raw SNR$=0$ (dB), we get a close performance to that of the trained network for each raw SNR. This indicates that in practice, one does not require the knowledge of the input raw SNR to achieve near optimal performance for a fixed BA duration. We also, observe a similar phenomena when fixing the raw SNR to zero (dB) and looking at the expected beamwidth for different BA duration. As shown in Fig.~\ref{fig:gen_ba}, by only training the network for $b=4$, we get near optimized performances for all $b\in[6]$ when raw SNR$=0$ (dB). These two observations suggest that training the network for a specific raw SNR and/or BA duration might be enough to achieve a near optimized performance which in turn reduces the complexity of using DNN based BA in practice. For our last simulation, we fix the training BA duration and compare the resulted expected beamwidth of our approach with the algorithm in \cite{chiu2019active} for different BA duration and raw SNRs. As shown in Fig.~\ref{fig:fb}, to outperform the state-of-the-art, training the network for only $b=4$ is enough.

\section{Conclusion}
In this paper, we have investigated the single-user interactive beam alignment problem in uplink systems where the objective is to minimize the expected width of the beam allocated to the user for data transmission. This allocated beam should include the AoA of the user with certain probability.  
We have formulated this problem as an end-to-end optimization of a DNN using an RNN. We have investigated effects of different loss functions and shown that the performance in terms of MMSE and expected beamwidth at high raw SNRs is close to that of the optimal BA when there is no noise for the case of uniform prior. Furthermore, we have observed that the proposed DNN based BA outperforms the state-of-the-art in all raw SNR regimes and different BA duration considering different priors. In this work, We have only considered perfect beams with the constraint that they should be contiguous. Adding more practical constraints is part of an ongoing work and is left for future publications.

\bibliographystyle{IEEEtran}
\bibliography{bbl}

\begin{thebibliography}{10}
\providecommand{\url}[1]{#1}
\csname url@samestyle\endcsname
\providecommand{\newblock}{\relax}
\providecommand{\bibinfo}[2]{#2}
\providecommand{\BIBentrySTDinterwordspacing}{\spaceskip=0pt\relax}
\providecommand{\BIBentryALTinterwordstretchfactor}{4}
\providecommand{\BIBentryALTinterwordspacing}{\spaceskip=\fontdimen2\font plus
\BIBentryALTinterwordstretchfactor\fontdimen3\font minus
  \fontdimen4\font\relax}
\providecommand{\BIBforeignlanguage}[2]{{%
\expandafter\ifx\csname l@#1\endcsname\relax
\typeout{** WARNING: IEEEtran.bst: No hyphenation pattern has been}%
\typeout{** loaded for the language `#1'. Using the pattern for}%
\typeout{** the default language instead.}%
\else
\language=\csname l@#1\endcsname
\fi
#2}}
\providecommand{\BIBdecl}{\relax}
\BIBdecl

\bibitem{mmWave-survey-nyu}
S.~Rangan, T.~S. Rappaport, and E.~Erkip, ``Millimeter-wave cellular wireless
  networks: Potentials and challenges,'' \emph{Proc. of the IEEE}, 2014.

\bibitem{rappaport2019wireless}
T.~S. Rappaport, Y.~Xing, O.~Kanhere, S.~Ju, A.~Madanayake, S.~Mandal,
  A.~Alkhateeb, and G.~C. Trichopoulos, ``Wireless communications and
  applications above 100 {GHz}: Opportunities and challenges for {6G} and
  beyond,'' \emph{IEEE Access}, 2019.

\bibitem{kutty2016beamforming}
S.~Kutty and D.~Sen, ``Beamforming for millimeter wave communications: An
  inclusive survey,'' \emph{IEEE Comms. Surveys \& Tutorials}, 2016.

\bibitem{akdeniz2014millimeter}
M.~R. Akdeniz, Y.~Liu, M.~K. Samimi, S.~Sun, S.~Rangan, T.~S. Rappaport, and
  E.~Erkip, ``Millimeter wave channel modeling and cellular capacity
  evaluation,'' \emph{IEEE JSAC}, 2014.

\bibitem{xing2018propagation}
Y.~Xing and T.~S. Rappaport, ``Propagation measurement system and approach at
  140 {GHz}-moving to {6G} and above 100 {GHz},'' in \emph{IEEE GLOBECOM},
  2018.

\bibitem{giordani2018tutorial}
M.~Giordani, M.~Polese, A.~Roy, D.~Castor, and M.~Zorzi, ``A tutorial on beam
  management for 3{GPP} {NR} at {mmWave} frequencies,'' \emph{IEEE Comms.
  Surveys \& Tutorials}, 2018.

\bibitem{barati2016initial}
C.~N. Barati, S.~A. Hosseini, M.~Mezzavilla, T.~Korakis, S.~S. Panwar,
  S.~Rangan, and M.~Zorzi, ``Initial access in millimeter wave cellular
  systems,'' \emph{IEEE TWC}, 2016.

\bibitem{giordani2016comparative}
M.~Giordani, M.~Mezzavilla, C.~N. Barati, S.~Rangan, and M.~Zorzi,
  ``Comparative analysis of initial access techniques in {5G} {mmWave} cellular
  networks,'' in \emph{IEEE CISS}, 2016.

\bibitem{Shah1906:Robust}
S.~Shahsavari, M.~Khojastepour, and E.~Erkip, ``Robust beam tracking and data
  communication in millimeter wave mobile networks,'' in \emph{Int. Symp. on
  Modeling and Opt. in Mobile, Ad Hoc, and Wireless Networks}, 2019.

\bibitem{khalili2020optimal}
A.~Khalili, S.~Shahsavari, M.~A.~A. Khojastepour, and E.~Erkip, ``On optimal
  multi-user beam alignment in millimeter wave wireless systems,'' \emph{IEEE
  ISIT}, 2020.

\bibitem{khalili2021single}
------, ``On single-user interactive beam alignment in millimeter wave systems:
  Impact of feedback delay,'' \emph{arXiv preprint arXiv:2102.02413}, 2021.

\bibitem{michelusi2018optimal}
N.~Michelusi and M.~Hussain, ``Optimal beam-sweeping and communication in
  mobile millimeter-wave networks,'' in \emph{IEEE ICC}, 2018.

\bibitem{hornik1989multilayer}
K.~Hornik, M.~Stinchcombe, H.~White \emph{et~al.}, ``Multilayer feedforward
  networks are universal approximators.'' \emph{Neural networks}, 1989.

\bibitem{hussain2017throughput}
M.~Hussain and N.~Michelusi, ``Throughput optimal beam alignment in millimeter
  wave networks,'' in \emph{IEEE ITA}, 2017.

\bibitem{khosravi2019efficient}
S.~Khosravi, H.~S. Ghadikolaei, and M.~Petrova, ``Efficient beamforming for
  mobile {mmWave} networks,'' \emph{arXiv preprint arXiv:1912.12118}, 2019.

\bibitem{chiu2019active}
S.-E. Chiu, N.~Ronquillo, and T.~Javidi, ``Active learning and {CSI}
  acquisition for {mmWave} initial alignment,'' \emph{IEEE JSAC}, 2019.

\bibitem{shabara2018linear}
Y.~Shabara, C.~E. Koksal, and E.~Ekici, ``Linear block coding for efficient
  beam discovery in millimeter wave communication networks,'' in \emph{IEEE
  INFOCOM}, 2018.

\bibitem{klautau20185g}
A.~Klautau, P.~Batista, N.~Gonz{\'a}lez-Prelcic, Y.~Wang, and R.~W. Heath,
  ``{5G} {MIMO} data for machine learning: Application to beam-selection using
  deep learning,'' in \emph{IEEE ITA}, 2018.

\bibitem{Song2019}
X.~{Song}, S.~{Haghighatshoar}, and G.~{Caire}, ``Efficient beam alignment for
  millimeter wave single-carrier systems with hybrid {MIMO} transceivers,''
  \emph{IEEE TWC}, 2019.

\bibitem{anton2019learning}
C.~Ant{\'o}n-Haro and X.~Mestre, ``Learning and data-driven beam selection for
  mmwave communications: An angle of arrival-based approach,'' \emph{IEEE
  Access}, 2019.

\bibitem{ma2020machine}
W.~Ma, C.~Qi, and G.~Y. Li, ``Machine learning for beam alignment in millimeter
  wave massive mimo,'' \emph{IEEE Wireless Comms. Letters}, 2020.

\bibitem{heng2019machine}
Y.~Heng and J.~G. Andrews, ``Machine learning-assisted beam alignment for
  mmwave systems,'' in \emph{IEEE GLOBECOM}, 2019.

\bibitem{wang2018mmwave}
Y.~Wang, M.~Narasimha, and R.~W. Heath, ``Mmwave beam prediction with
  situational awareness: A machine learning approach,'' in \emph{IEEE SPAWC},
  2018.

\bibitem{alkhateeb2018deep}
A.~Alkhateeb, S.~Alex, P.~Varkey, Y.~Li, Q.~Qu, and D.~Tujkovic, ``Deep
  learning coordinated beamforming for highly-mobile millimeter wave systems,''
  \emph{IEEE Access}, 2018.

\bibitem{sohrabi2020deep}
F.~Sohrabi, Z.~Chen, and W.~Yu, ``Deep active learning approach to adaptive
  beamforming for mmwave initial alignment,'' \emph{arXiv preprint
  arXiv:2012.13607}, 2020.

\bibitem{nguyen2018comparing}
S.~L. Nguyen, J.~Jarvelainen, A.~Karttunen, K.~Haneda, and J.~Putkonen,
  ``Comparing radio propagation channels between 28 and 140 {GHz} bands in a
  shopping mall,'' in \emph{IET EuCAP}, 2018.

\bibitem{ramanathan2001performance}
R.~Ramanathan, ``On the performance of {Ad} {Hoc} networks with beamforming
  antennas,'' in \emph{Proc. of ACM Int. Symp. on Mobile Ad Hoc Networking \&
  Computing}, 2001.

\bibitem{bai2015coverage}
T.~Bai and R.~W. Heath, ``Coverage and rate analysis for millimeter-wave
  cellular networks,'' \emph{IEEE TWC}, 2015.

\bibitem{bertsekas1995dynamic}
D.~P. Bertsekas, \emph{Dynamic Programming and Optimal Control}.\hskip 1em plus
  0.5em minus 0.4em\relax Athena scientific Belmont, MA, 1995, vol.~1, no.~2.

\bibitem{Alkhateeb2014}
A.~{Alkhateeb}, O.~{El Ayach}, G.~{Leus}, and R.~W. {Heath}, ``Channel
  estimation and hybrid precoding for millimeter wave cellular systems,''
  \emph{IEEE JSSP}, 2014.

\end{thebibliography}

\end{document}